\documentclass[10pt,twoside]{article}

\usepackage{times}
\usepackage[utf8]{inputenc}
\usepackage[T1]{fontenc}
\usepackage{graphicx}
\usepackage{algorithm}
\usepackage{adjustbox}%

\usepackage{siunitx}
\usepackage{hyperref}
\usepackage{cleveref}
\usepackage{coria-taln2025}
\usepackage[french]{babel} 

\title{Backtesting Sentiment Signals for Trading: Evaluating the Viability of Alpha Generation from Sentiment Analysis}

\author{Elvys Linhares Pontes\up{1}, Carlos-Emiliano González-Gallardo\up{2}, Georgeta Bordea\up{3}, José G. Moreno\up{4}, Mohamed Ben Jannet\up{1}, Yuxuan Zhao\up{1} and Antoine Doucet\up{3}\\
  {\small
    (1) Trading Central Labs - Trading Central, Paris, France \\
    (2) LIFAT, University of Tours, Tours, France \\
    (3) L3i, University of La Rochelle, La Rochelle, France \\
    (4) IRIT, University of Toulouse, Toulouse, France \\
    \texttt{
      elvys.linharespontes@tradingcentral.com \\ 
    }
  }
}

\begin{document}
\maketitle

\resume{
L'analyse de sentiment, largement utilisée dans les avis de produits, influence également les marchés financiers en affectant les prix des actifs à travers les microblogs et les articles de presse. Bien que la recherche sur la finance basée sur le sentiment soit abondante, de nombreuses études se concentrent sur la classification au niveau des phrases, négligeant son application pratique dans le trading. Cette étude comble cette lacune en évaluant des stratégies de trading basées sur le sentiment pour générer un alpha positif. Nous réalisons une analyse de backtesting en utilisant des prédictions de sentiment de trois modèles (deux basés sur la classification et un basé sur la régression) appliqués aux articles de presse concernant les actions du Dow Jones 30, en les comparant à la stgonzalezgallardo@univ-tours.frratégie de référence Buy\&Hold. Les résultats montrent que tous les modèles ont généré des rendements positifs, le modèle de régression enregistrant le rendement le plus élevé de 50,63\% sur 28 mois, surpassant ainsi la stratégie Buy\&Hold. Cela souligne le potentiel de l'analyse de sentiment pour affiner les stratégies d'investissement et améliorer la prise de décisions financières.
}

\abstract{}{
\vspace{-0.6cm}
Sentiment analysis, widely used in product reviews, also impacts financial markets by influencing asset prices through microblogs and news articles. Despite research in sentiment-driven finance, many studies focus on sentence-level classification, overlooking its practical application in trading. This study bridges that gap by evaluating sentiment-based trading strategies for generating positive alpha. We conduct a backtesting analysis using sentiment predictions from three models (two classification and one regression) applied to news articles on Dow Jones 30 stocks, comparing them to the benchmark Buy\&Hold strategy. Results show all models produced positive returns, with the regression model achieving the highest return of 50.63\% over 28 months, outperforming the benchmark Buy\&Hold strategy. This highlights the potential of sentiment in enhancing investment strategies and financial decision-making.
}

\motsClefs
  {Classification du sentiment; Stratégie de trading basée sur les sentiments ; Backtesting financier ; Prévision des rendements des prix}
  {Sentiment classification; Sentiment-based trading strategy; Financial backtesting; Forecasting price return}


\section{Introduction}

Sentiment analysis is a natural language processing task that has gained significant attention in recent years, with application domains such as product reviews, brand perception, and public sentiment towards political events or social issues. However, the influence of sentiments and opinions is not limited to consumer preferences and public discourse, reaching into the complex dynamics of financial markets. A growing body of research \citep{jin2020stock,oliveira2021investment} shows that sentiments and opinions play a pivotal role in shaping the behaviors and outcomes of market entities. Sentiments and opinions expressed in financial microblogs and news articles can significantly impact market dynamics, often leading to abrupt fluctuations in asset prices. Recent works address the challenge of sentiment analysis of financial microblogs and news to predict market reactions \citep{Araci2019FinBERTFS,Wan:2021}.  Therefore, integrating sentiment analysis into financial market analysis can potentially improve investment strategies and risk management. At the same time, it is essential to acknowledge that news outlets can also introduce distortions, resulting in producers' and consumers' biases, noisy information, and speculative sentiments \citep{NBERw29344}.

Most of the literature in finance focuses on sentiment analysis at the level of sentences or news articles~\cite{cortis-etal-2017-semeval,Mishev2020}. Sentiment analysis can be conceptually framed as either a classification or regression task, each offering varying degrees of granularity in sentiment interpretation. 
Few works in the existing literature make a critical evaluation of the practical applicability of sentiment analysis in financial trading~\cite{lou2023stockmarketsentimentclassification,Kargarzadeh2024}. An effective method to assess the impact of sentiment analysis on financial markets is to develop trading strategies exclusively based on sentiment data. These strategies operate according to specific rules to determine the optimal times to execute buy or sell orders for an asset. By analyzing the outcomes of these sentiment-driven trading strategies, researchers can evaluate whether the sentiment data provided by a model contributes to generating profits or incurs losses. Such evaluations are instrumental in understanding the real-world practicality and effectiveness of sentiment analysis in financial trading.

Despite recent advances in sentiment model prediction, a persistent challenge remains in evaluating the practical efficacy of the predicted sentiment scores. Although previous studies have focused mainly on assessing the accuracy of a model in predicting sentiment compared to human evaluations~\cite{cortis-etal-2017-semeval,Mishev2020} or combining sentiment/emotion with macroeconomic and technical data to forecast price returns~\cite{8113051,Kargarzadeh2024}, our study adopts a distinctive approach.
In this work, our objective is to comprehensively evaluate the effectiveness of solo sentiment analysis strategies by examining their ability to produce actionable sentiment scores that inform trading strategies. This includes assessing their impact on trading performance and determining whether they can generate a positive alpha.

Financial backtesting is a critical tool that simulates how a particular strategy would have performed in historical market conditions. This approach offers valuable insights into risk management, profitability, and overall strategy viability.
Therefore, we propose a comparative backtesting trading strategy analysis that leverages sentiment predictions from three distinct sentiment recognition models. 
These models, which differ in their complexity, are used to forecast daily price return variations.
We compared these methods with a benchmark strategy consisting of a \textit{Buy\&Hold} approach for the assets over the backtesting period. The backtesting involved determining daily sentiment trends from news articles published between January 1, 2020, and April 30, 2022, focusing on content relevant to the Dow Jones 30 stocks and generating daily signals to \textit{buy} or \textit{sell} an asset. In our analysis, all sentiment models generated positive cumulative returns at the end of the backtesting period. As expected, the regression model produced the highest cumulative return of 50.63\%, which is 87.8\% higher than the returns from the classification methods and the benchmark, thus generating a considerable alpha of 23.67\%.
This outcome highlights the potential of sentiment analysis-based trading systems as compelling investment strategies in financial markets. Moreover, it highlights their versatility and potential for integration with additional information, such as macroeconomic indicators and technical data, to further enhance trading performance.

The rest of this paper is structured as follows: Section~\ref{sec:related} reviews related work on the application of sentiment analysis in backtesting and its impact on financial markets. Section~\ref{sec:case_description} describes our backtesting procedure, including details on data collection, selected sentiment models, sentiment aggregation, and trading decision processes. Section~\ref{sec:results} presents the experiments conducted and discusses the results of our analysis. Finally, Section~\ref{sec:conclusion} concludes the paper with key insights and final remarks on our case study.

\section{Related Work}
\label{sec:related}

Sentiment analysis has been widely applied across diverse domains, including assessing product quality via user feedback or measuring public sentiment about a stock on social networks. However, relatively few studies have explored the application of sentiment analysis specifically in the context of trading evaluation.

\citet{kazemian2016evaluating} incorporated sentiment into a trading strategy using an SVM-based sentiment classifier. Their approach involved acquiring stocks when news articles were classified as having a positive sentiment and selling them when sentiment was negative. While their method demonstrated a strong correlation between sentiment signals and trading actions, it was sensitive to daily variations, mainly when multiple articles related to the same company were analyzed on the same day.
\citet{lou2023stockmarketsentimentclassification} extended this idea by examining the impact of human emotions on quantitative trading. By fine-tuning the BERT model on user comments from East Money and integrating it with the Alpha191 model, Lou achieved a 73.8\% improvement in return rates compared to a baseline strategy.

\citet{Bird:2023} further explored the integration of emotions into trading strategies, focusing on the timing of stock purchases, short sales, and reversals. Their approach involved acquiring stocks shortly after exceptionally positive earnings news when positive emotions were subdued and reversing the position based on either an increase in positive emotions or the end of the holding period. For short positions, they employed a similar method following extremely negative earnings news. Their findings underscored the significant influence of investor emotions on individual investment decisions, suggesting that price distortions are corrected only when emotions shift.

\citet{8113051} analyzed the relationship between stock price movements and social media sentiment in China, leveraging data from platforms such as microblogs, chat rooms, and forums. Their study revealed a strong correlation between chat room sentiment and stock price fluctuations. By developing a trading strategy based on chat room sentiment, they achieved a portfolio return of 19.54\% over seven months, significantly outperforming a passive buy-and-hold approach.

\citet{Kargarzadeh2024} presented a trading strategy combining large language models (LLMs) with macroeconomic and technical indicators to enhance stock return predictions, focusing on small-cap stocks from the Russell 2000 Index. By incorporating sentiment analysis from financial news using GPT-4 and applying a decay function for news impact, their strategy achieved exceptional performance, with Sharpe ratios of 3.64 and 5.10 in 2022 and 2023, respectively, even after accounting for transaction costs.

Focused on the bias in the backtesting, \citet{glasserman2023assessinglookaheadbiasstock} examined backtesting trading strategies based on sentiment derived from LLMs. They identified two key biases—look-ahead and distraction effects—and demonstrated that anonymizing financial news headlines by removing company identifiers improved trading performance. Their findings suggest that the distraction effect has a more significant impact than look-ahead bias, particularly for larger companies, and their proposed anonymization technique offers a novel method for reducing bias in backtesting.

Our work focuses on developing a trading system based solely on sentiment analysis over an extended period of more than two years. While~\citet{Bird:2023} focused on the emotions of earning news to generate their signals, we evaluate three distinct sentiment approaches to create trading systems that derive daily trading signals from major financial news providers (all types of news) on the Dow Jones 30 stocks. The following section outlines the backtesting methodology and protocol used in our study.

\section{Backtesting Description}
\label{sec:case_description}
\subsection{Sentiment-based Trading Strategy}


Our primary objective in this study is to quantify the direct impact of news sentiment on a trading system, employing news sentiment as the sole criterion to determine market entry. To develop our trading strategy, we embarked on a comprehensive analysis of assets within the Dow Jones 30 index, covering a period of more than two years. The proposed strategy relies on predicting the sentiment for each of the entities mentioned in the news articles to guide our investment decisions.

We start by meticulously scrutinizing articles from trusted sources that reference any assets listed in the Dow Jones 30 index (\textbf{Data Collection}).
Then, using various sentiment analysis methodologies, we assess the sentiment associated with each asset mentioned in the selected articles, allowing us to accurately gauge market sentiment (\textbf{Sentiment Analysis Model}). 
Before the market opens each day, we aggregate the sentiments expressed in articles published between the previous and the current market open. This aggregation results in a unique sentiment score for each asset, which serves as a critical input for our decision-making process (\textbf{Sentiment Aggregation}).
Depending on this sentiment score for each asset, we generate \textit{buy}, \textit{neutral}, or \textit{sell} orders. These orders are executed at the market's opening price, ensuring that our strategy aligns with current market conditions. This process is repeated daily throughout the backtesting period analyzed, allowing us to adapt to the evolving market dynamics (\textbf{Trading Decisions}). Finally, we meticulously evaluate the performance of this sentiment analysis approach.
In the subsequent subsections, we will delve into the specifics of each stage within this strategic pipeline.

\subsubsection{Data Collection}
\label{sec:datacollection}

We conducted an extensive analysis of articles collected from multiple reputable financial and economic providers and sources, encompassing the backtesting period from January 1st, 2020, to April 30th, 2022\footnote{Further details about this corpus can be consulted at Appendix~\ref{sc:corpus_stats}.}. 
Although the backtesting period is relatively limited, it is specifically selected to encompass the most significant decline in the history of the Dow Jones 30 index and the subsequent recovery period.
Specifically, the article selection process involved curating content from a variety of respected financial sources, including Accesswire\footnote{\url{https://www.accesswire.com}}, 
Benzinga\footnote{\url{https://www.benzinga.com/}}, Infocast\footnote{\url{https://infocastinc.com}}, 
Informa\footnote{\url{https://www.informa.com}}, 
MT Newswires\footnote{\url{https://www.mtnewswires.com}}, 
Reuters\footnote{\url{https://www.reuters.com}},
and Seeking Alpha\footnote{\url{https://www.seekingalpha.com}}. 
We also used a web crawler to retrieve other reliable financial sources such as American Banking News\footnote{\url{https://www.americanbankingnews.com}} 
and Yahoo Finance\footnote{\url{https://finance.yahoo.com/}}.

All of these articles are raw and contain no meta-information. To recognize and map company names of the Dow Jones 30 index on these texts, we employed an entity recognizer system based on a BERT-based model integrated with a CRF~\cite{abs-1909-10649}. 
Our primary objective was to identify and isolate sentences within the articles that mentioned specific assets. This allowed us to focus sentiment analysis on only the portions of the text directly relevant to each asset.
Although sentiment is extracted at the sentence level, we calculate sentiment once per article for each asset by aggregating the sentiment scores of all sentences within that article where the asset is mentioned. This produces a single sentiment score per asset per article, thereby maintaining the article as the unit of analysis.

\subsubsection{Sentiment Analysis Models}

To assess the influence of various sentiment analysis methods on a trading strategy, we build upon the analytical framework introduced by \citet{Mishev2020}, which compared multiple sentiment analysis approaches in the financial domain. Our study employs three sentiment analysis models based on BERT architectures to predict the sentiment of financial news articles and evaluate their impact in the backtesting analysis.
The first two models, FinBERT~\cite{Araci2019FinBERTFS} and DualGCN~\cite{li2021dual}, approach sentiment analysis as a classification task, categorizing sentiment into three discrete classes: \textit{negative}, \textit{neutral}, and \textit{positive}. To provide a comparison with a regression-based approach, we implemented the RoBERTa+Transfs model~\cite{roberta-transformer}, which predicts sentiment as a continuous value within the range of -1.0 to +1.0 and outperformed other state-of-the-art sentiment analysis systems on financial data.

The FinBERT model\footnote{\url{https://huggingface.co/ProsusAI/finbert}} served as the foundational pre-trained model for our classification efforts.
Regarding the DualGCN model, we trained it on the aggregated financial dataset composed of 4,846 sentences from the Financial Phrase Bank v.1.0 and 1,633 news headlines from the SemEval-2017 task \#5~\cite{cortis-etal-2017-semeval}. 
Concerning the RoBERTa+Transfs model, we followed the same procedure to train the model as described in \citet{roberta-transformer}\footnote{Further details and training procedures can be consulted at Appendix~\ref{sc:train_models}.}.



To position these methods, we propose a baseline consisting of the straightforward \textit{Buy\&Hold} strategy, which serves as the Dow Jones 30 index benchmark. This strategy involves deploying all available capital on the first day by purchasing an equal amount of shares for each stock and retaining them for the entire duration of the backtesting period. All stocks held in the portfolio are sold on the last day of the backtesting period.

\subsubsection{Sentiment Aggregation}

To establish distinct sentiment measures for individual assets and facilitate the formulation of effective trading strategies, we aggregate sentiment scores derived from articles published from the previous market open day to the current market open day (i.e., 9:30 a.m. local time).
Each day before the markets open and for each asset, we average the sentiment present in articles published between the previous market open day and the current market open day. Aggregating sentiment in this manner allows us to derive a distinct sentiment score for each asset, which informs our trading strategy. This aggregation employs a scale from 0 to 100, where 0 represents a highly negative sentiment, and 100 signifies an exceptionally positive sentiment.

As the FinBERT and DualGCN predictions are only labels, we replace the \textit{negative}, \textit{neutral}, and \textit{positive} labels by the values $-1$, 0, and $+1$, respectively, for the calculation of the sentiment aggregation. Regarding RoBERTa+Transfs, we use the predicted scores directly in the sentiment aggregation. Then, we transform the average sentiment of these articles for each method into values within the sentiment aggregation range from 0 to 100.

\subsubsection{Trading Decisions}
 
Before the market's opening, we formulate specific order lists for each asset based on its respective aggregated sentiment scores. Our ordering criteria are as follows:

If an asset's sentiment score exceeds \textrm{BUY\_SIGNAL}, we generate a \textit{Buy} order; if it falls within the range of \textrm{SELL\_SIGNAL} to \textrm{BUY\_SIGNAL}, we issue a \textit{Neutral} order, and if the sentiment score is below \textrm{SELL\_SIGNAL}, we initiate a \textit{Sell} order.
We have chosen to employ an \textit{equal-value} order strategy, meaning each buy/sell order involves a fixed amount of \textrm{ORDER\_VALUE}.
We then execute these orders sequentially on a daily basis. Initially, we compile the \textit{Buy} list, reviewing each instrument therein. If an instrument is already part of our portfolio, no action is taken; otherwise, we execute a purchase. Moving on, we address the \textit{Neutral} list by closing positions associated with instruments found within it. 
Finally, we evaluate the instruments listed in the \textit{Sell} category. If an instrument is already in our portfolio, we do not take action. However, if it is not, we engage in short-selling.

In our strategy, we implemented the \textit{equal-value} approach, which means that regardless of the number of signals triggered, we consistently buy and sell a fixed amount of money's worth of stock. For instance, if there are 3 \textit{Buy} and 2 \textit{Sell} orders, we will buy \textrm{ORDER\_VALUE} per each signal stock and sell \textrm{ORDER\_VALUE} per each sell stock. Only in the best-case scenario will our money be 100\% efficient, but most of the time, our money will be idle. 
The orders are executed at the market open, using the opening price as the reference value.

It is important to note that in our analysis we assume commission fees of 0.05\% of the traded value. Our initial capital is set at \$300,000 (since we have 30 stocks in our list and each order (\textrm{ORDER\_VALUE}) is set to a fixed value of \$10,000), and our strategy operates with a daily trading frequency. 

To determine the optimal parameters for \textrm{SELL\_SIGNAL} and \textrm{BUY\_SIGNAL}, we performed comprehensive analyses to achieve stable transactions and consistent results. Beginning with the neutral value of 50, we systematically tested signal thresholds at 45 and 55, 40 and 60, and 35 and 65. Regrettably, the thresholds of 45 and 55 resulted in excessive signal activations, leading to a significant volume of transactions. In contrast, the thresholds of 40 and 60 exhibited a more balanced and stable transaction frequency, with the signals being activated appropriately over time. Subsequently, the thresholds of 35 and 65 produced a significantly lower transaction volume, accompanied by inadequate signal activations. Consequently, we established \textrm{SELL\_SIGNAL} and \textrm{BUY\_SIGNAL} at 40 and 60, respectively, as they demonstrated the most desirable balance between transaction stability and signal activation frequency.

\section{Evaluation and Discussion}
\label{sec:results}

\subsection{Evaluation metrics}

In the evaluation of financial backtesting, several key performance metrics are of significant relevance, providing valuable information on the effectiveness and risk associated with investment strategies. In this work, we selected metrics such as annual, annual compound, and annual cumulative returns; Calmar, Sharpe, and Sortino ratios; max drawdown (MDD); annual volatility; and 95\% daily value at risk (VaR)\footnote{Additional details about the metrics can be found at Appendix~\ref{sc:annex_metrics}.}. These metrics collectively offer a holistic view of a strategy's performance, risk, and resilience. Investors rely on these measures to assess the profitability and risk associated with their investments.

\subsection{Results and Discussion}

\begin{figure*}[!h]
    \centering
    \includegraphics[width=0.85\linewidth]{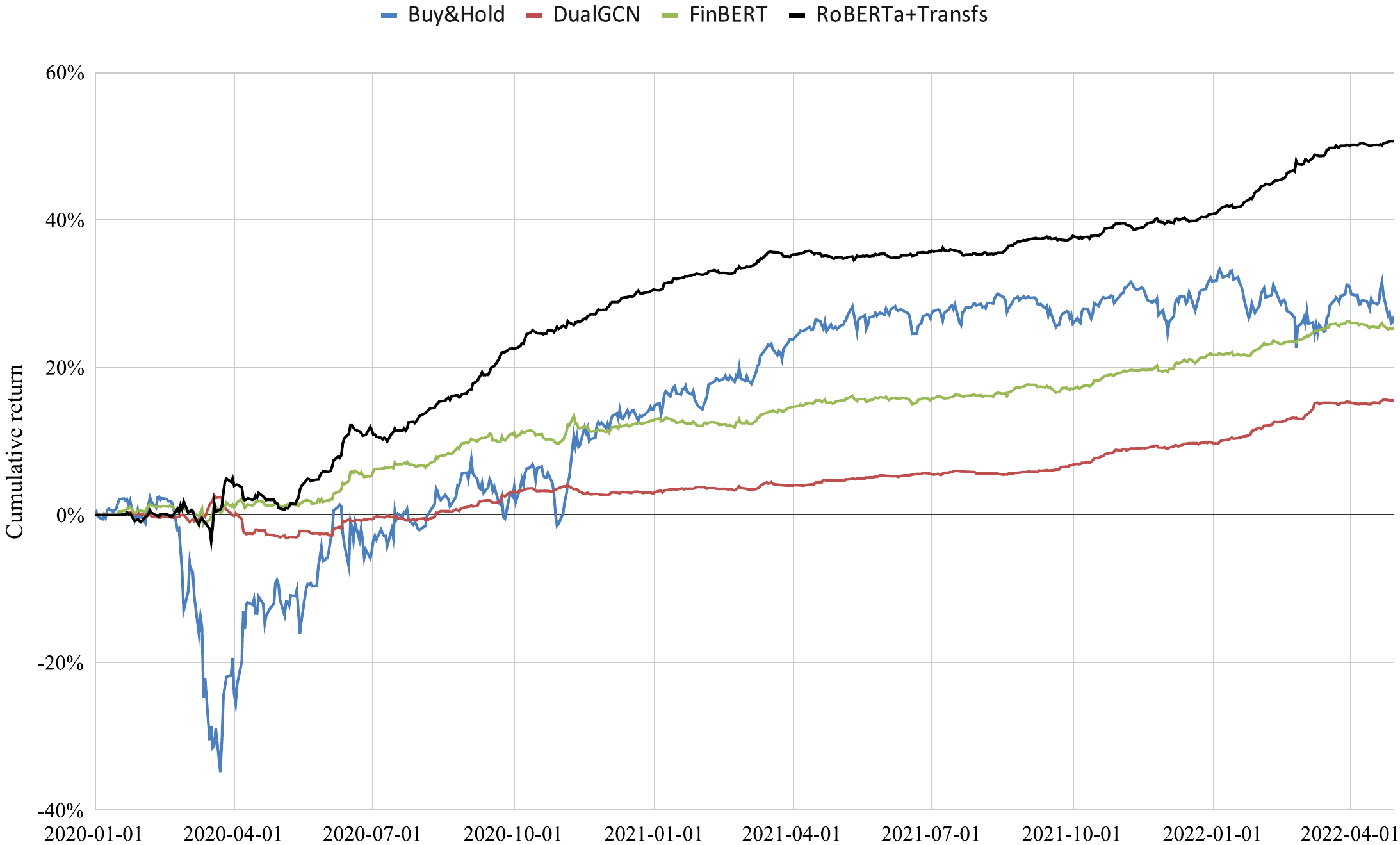}
    \caption{Cumulative return of all approaches over the backtesting period.}
    \label{fig:cumulative}
\end{figure*}

A detailed examination of annual, compound, and cumulative returns in Table~\ref{tb:backtest} reveals that the sentiment predictions generated by the FinBERT model outperform those produced by the DualGCN model when applied within the framework of our trading strategy. Interestingly, even though the baseline \textit{Buy\&Hold} strategy is more straightforward, it achieved a cumulative return of 26.96\%, exceeding both the DualGCN and FinBERT approaches.
It is also worth noting that sentiment predictions generated by the RoBERTa+Transfs model yield superior returns compared to all other methods (Figure \ref{fig:cumulative}). This advantage can be attributed to the fact that RoBERTa+Transfs predicts sentiment scores as floating-point values, rather than discrete classes, which results in a more precise sentiment score that is better suited for our application.

\begin{table*}[ht]
\centering
\begin{adjustbox}{width=\linewidth}
\begin{tabular}{|l|c|c|c|c|}
\hline
\textbf{Metrics}  & \textbf{Buy\&Hold} & \textbf{DualGCN}  & \textbf{FinBERT}  & \textbf{RoBERTa+Transfs}    \\\hline
Annual Return     & 9.12\%    & 6.58\%   & 10.91\%  & \textbf{23.15\%}  \\ \hline
Annual Compound Return       & 23.44\%   & 16.62\%  & 28.39\%  & \textbf{65.26\%}  \\\hline
Annual Cumulative Return       & 26.96\%   & 15.49\%  & 25.23\%  & \textbf{50.63\%}  \\\hline
Calmar Ratio      & 0.28      & 1.16     & 3.97     & \textbf{4.39} \\\hline
Sharpe Ratio      & 0.38      & 1.92     & 2.31     & \textbf{3.98}     \\\hline
Sortino Ratio     & 0.69      & 3.37     & 3.64     & \textbf{7.06} \\\hline
MDD      & -32.38\%  & -5.67\%  & \textbf{-2.75\%}  & -5.27\%  \\\hline
Annual Volatility & 22.05\%   & \textbf{3.01\%}   & 4.38\%   & 5.61\%   \\\hline
95\% Daily VaR  & 1.89\%    & \textbf{0.25\%}   & 0.44\%   & 0.56\%   \\ \hline
\end{tabular}
\end{adjustbox}
\caption{\label{tb:backtest} Backtesting results over the period between January 1st, 2020, and April 30th, 2022. Text in \textbf{bold} indicates the best strategy for each metric.}
\end{table*}

\begin{figure*}[!h]
    \centering
    \includegraphics[width=0.85\linewidth]{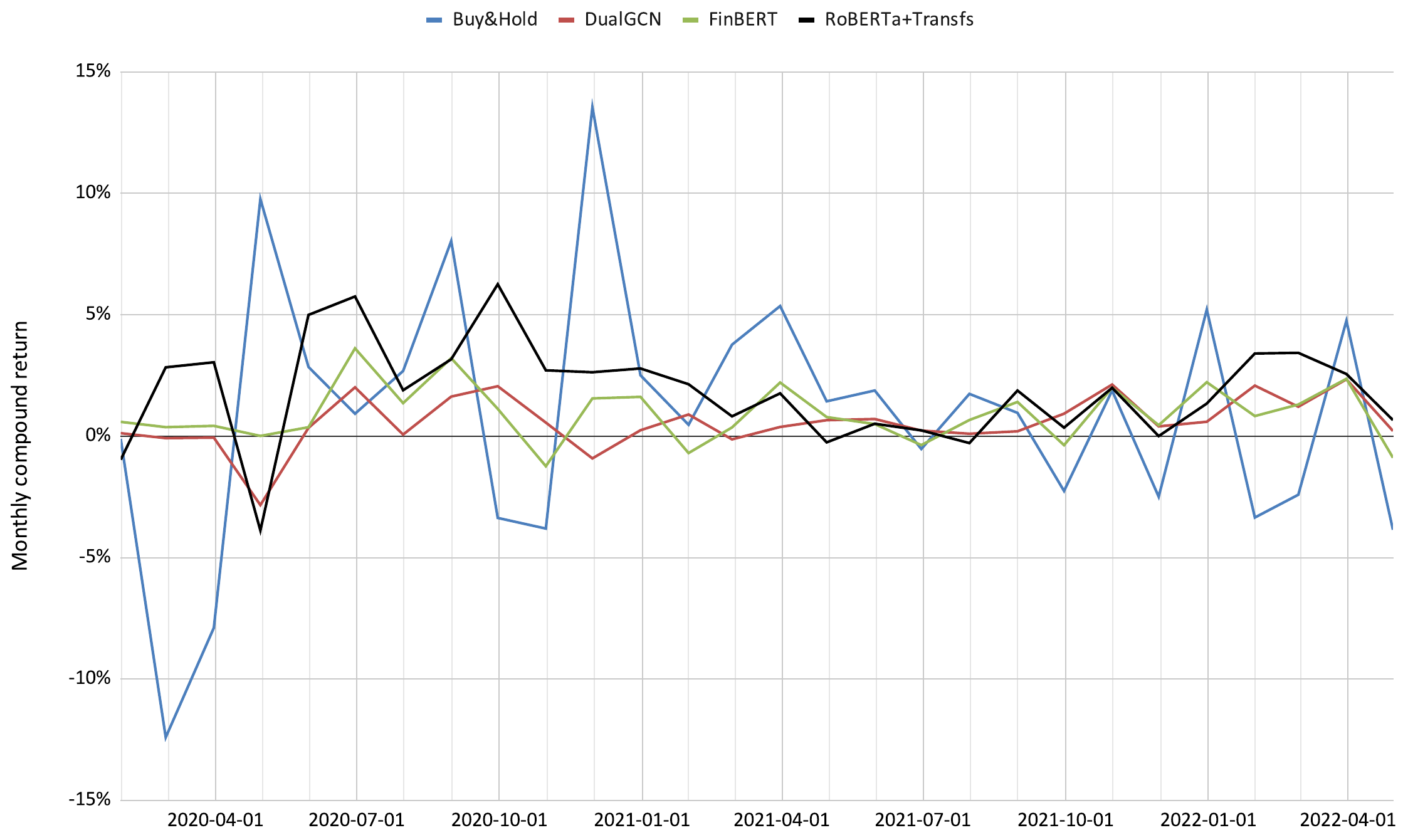}
    \caption{Monthly compound return of all approaches over the backtesting period.}
    \label{fig:monthly-compound}
\end{figure*}

An in-depth examination of Table \ref{tb:backtest} further reveals that the RoBERTa+Transfs strategy consistently achieves the highest scores in terms of the Calmar, Sharpe, and Sortino ratios. These metrics indicate that the strategy based on RoBERTa+Transfs exhibits the best risk-adjusted performance among all the approaches considered.
In particular, FinBERT achieves the best max drawdown, which indicates the smallest losses among all methods. RoBERTa+Transfs secures the second position with a maximum drawdown of -5.27\%, which is considered acceptable given the cumulative return achieved.

Analyzing annual volatility, we find that the strategies built on the RoBERTa+Transfs model tend to exhibit high volatility, while those constructed on DualGCN demonstrate the lowest volatility. Nevertheless, it is essential to emphasize that even the 5.61\% annual volatility for RoBERTa+Transfs remains a promising result, and the high return well justifies the additional risk it presents. This same trend extends to 95\% Daily VaR, with DualGCN achieving the best results, and RoBERTa+Transfs ranking third.

The \textit{Buy\&Hold} strategy records higher annual volatility and daily VaR compared to other approaches, mainly due to the utilization of an \textit{equal-value} strategy in our backtesting, as opposed to an \textit{equal-weight} strategy. Indeed, the \textit{equal-weight} strategy gives every asset an equal share of total capital, which can lead to higher risk and volatility because it does not account for variations in individual asset characteristics. On the other hand, the \textit{equal-value} strategy allocates the same dollar amount to each asset, leading to a more balanced and diversified portfolio, which can help reduce risk and volatility.

Examining the monthly return shown in Figure~\ref{fig:monthly-compound}, it is clear that DualGCN, FinBERT, and RoBERTa+Transfs mainly generated positive returns throughout the entire backtesting period. In contrast, the baseline \textit{Buy\&Hold} strategy encountered several months of negative returns, especially during the initial lockdown phase triggered by the Covid-19 pandemic, resulting in a cumulative return of -34.87\% during that period. Amid the severe economic impacts of the pandemic, many governments began announcing unprecedented economic rescue packages starting in mid-March. These measures have helped restore investor confidence, as reflected in the recoveries seen across most stock market~\citep{SEVEN2021101349,brookings_recovery_2021}. As a result, the baseline strategy yielded a positive return in 2020, despite the downturn in the first half of the year, and achieved the highest return among all approaches in 2021.
The downturn in the baseline's performance in 2022 can be primarily attributed to two key factors: first, the surge in inflation in December 2021, and second, the Federal Reserve's multiple interest rate hikes during the first half of 2022. Furthermore, the escalation of the Ukraine-Russia conflict, which began in February 2022, exacerbated market volatility. These events underscore the limitations of the \textit{Buy\&Hold} strategy in effectively responding to major economic and geopolitical developments to mitigate investment risks. Despite these challenges, all other strategies successfully adapted to these events, demonstrating their ability to generate positive returns during this period.

Although achieving a better final return compared to FinBERT and DualGCN, the \textit{Buy\&Hold} strategy exhibited fluctuations in performance, with the most significant drop occurring during the initial lockdown period. In contrast, RoBERTa+Transfs consistently outperformed other approaches for most of the backtesting period and notably achieved the highest annual return in 2020 and 2022. Although there were instances in which the baseline outperformed RoBERTa+Transfs in isolated months, the latter consistently demonstrated higher returns over time. 

The annual compound returns for the strategies are as follows: In 2020, RoBERTa+Transfs achieved 35.26\%, FinBERT 13.47\%, Buy\&Hold 9.98\%, and DualGCN 2.86\%. In 2021, Buy\&Hold led with 18.24\%, followed by RoBERTa+Transfs at 10.73\%, FinBERT at 9.27\%, and DualGCN at 7.07\%. By 2022, returns declined across most strategies: RoBERTa+Transfs recorded 10.34\%, DualGCN 5.89\%, FinBERT 3.55\%, and Buy\&Hold -5.07\%. It is noteworthy that RoBERTa+Transfs achieved the highest return among the approaches in most cases, particularly excelling in 2020 and 2022. While securing the best return in 2021, the baseline strategy ranked as the second worst in 2020 and experienced negative returns in 2022, reducing its total compound return. These observations underscore the consistent and superior performance of RoBERTa+Transfs in various market conditions, positioning it as a robust choice for investment strategies.

Finally, we examined the sentiment-based models' alpha generation. Alpha represents the excess return of an investment or portfolio relative to a benchmark index. It is calculated as the difference between the portfolio's return and the benchmark's return. RoBERTa+Transfs achieved an impressive alpha of 23.67\%, while DualGCN and FinBERT recorded -11.47\% and -1.73\%, respectively.
When an investment strategy generates a positive alpha, it indicates that the investor has achieved higher returns than the broader market. 
The limitation of three classes reduced the sentiment analysis capacity of models, which affected the cumulative return, thus generating a negative alpha. However, the floating-based model generated a considerable positive alpha of 23.67\% with a lower risk than the benchmark (\textit{Buy\&Hold}). These results substantiate the utility of sentiment analysis within the financial market context, highlighting that sentiment data can be highly valuable to the effectiveness of trading strategies.

\section{Conclusion}
\label{sec:conclusion}

In this paper, we proposed an evaluation framework based on financial backtesting to measure the effectiveness of incorporating sentiment scores predicted by machine learning approaches from news articles into a trading strategy. 
The backtesting from January 1, 2020, to April 30, 2022, demonstrates that utilizing news sentiment significantly contributes to creating a profitable trading strategy with relatively low risk. Furthermore, we observed that, opposed to discrete classes, models predicting sentiment as floating-point values result in improved performance.
The RoBERTa+Transfs approach achieved the highest cumulative return $50.63\%$ which is $87.8\%$ higher than all other methods analyzed in this study. The significant gap between FinBERT and RoBERTa+Transfs is primarily attributed to FinBERT predicting only three possible classes, whereas RoBERTa+Transfs can estimate various degrees of positivity or negativity in an article. Moreover, the floating-based model generated a considerable alpha of 23,67\% with a lower risk than the benchmark (\textit{Buy\&Hold)}.

Future work will aim to conduct a more comprehensive comparison of trading strategies driven exclusively by news sentiment against hybrid approaches that integrate sentiment analysis with technical indicators and macroeconomic variables. The objective is to enhance the precision and effectiveness of daily order execution. This analysis will be strengthened by extending the backtesting period to cover at least 5 years, providing a more robust assessment of strategy performance across varying market cycles, including bull and bear phases. The analysis will also be expanded to include major indices such as the S\&P 500 and NASDAQ-100 to evaluate strategy robustness across different market environments.

\section*{Acknowledgments}

This work has been supported by the France Relance (ANR-21-PRRD-0010-01) project funded by the French National Research Agency (ANR).

\bibliographystyle{coria-taln2025}
\bibliography{biblio}
\nocite{TALN2015,LaigneletRioult09,LanglaisPatry07,SeretanWehrli07}

\appendix
\section{Appendix : Corpus statistics}
\label{sc:corpus_stats}

The corpus comprises a total of 1,332,446 news articles distributed across 28 major companies, with each company contributing a varying number of articles (Table~\ref{tb:corpus}). On average, each company has approximately 47,587 articles over the backtesting period. However, there is notable variation, ranging from as few as 4,440 articles for The Travelers Companies, Inc. to over 231,000 for Amazon.com, Inc. The articles contain on average a title length of 14 words and an average text length of 799 words.

\begin{table}[!h]
\centering
\begin{tabular}{|l|c|c|c|c|c|}
\hline
\multicolumn{1}{|c|}{\textbf{Company}} & \textbf{\#articles} & \textbf{\#title} & \textbf{\#text} & \textbf{\#vocab title} & \textbf{\#vocab text} \\ \hline
3M Company & 9772 & 14 & 989 & 10552 & 106749 \\ \hline
Amazon.com, Inc. & 231937 & 14 & 976 & 110619 & 1028454 \\ \hline
American Express Company & 15676 & 14 & 819 & 16392 & 147064 \\ \hline
Amgen Inc. & 14343 & 14 & 956 & 13990 & 143150 \\ \hline
Apple Inc & 223781 & 14 & 697 & 98533 & 865425 \\ \hline
Caterpillar Inc. & 13188 & 14 & 863 & 13419 & 135102 \\ \hline
Chevron Corporation & 30758 & 14 & 835 & 24707 & 226205 \\ \hline
Cisco Systems, Inc. & 27415 & 16 & 903 & 24291 & 217317 \\ \hline
Honeywell International Inc. & 16156 & 16 & 882 & 16419 & 148084 \\ \hline
Intel Corporation & 46425 & 14 & 778 & 36275 & 313935 \\ \hline
IBM Corporation & 31420 & 15 & 819 & 29820 & 276171 \\ \hline
Johnson \& Johnson & 51114 & 15 & 770 & 38691 & 332024 \\ \hline
McDonald's Corporation & 25758 & 14 & 749 & 26675 & 214226 \\ \hline
Merck \& Co., Inc. & 26316 & 16 & 951 & 23554 & 235094 \\ \hline
Microsoft Corporation & 138576 & 14 & 783 & 69969 & 701322 \\ \hline
NIKE, Inc. & 27770 & 14 & 761 & 26768 & 222125 \\ \hline
Salesforce, Inc. & 20037 & 14 & 777 & 20421 & 160254 \\ \hline
The Boeing Company & 59354 & 14 & 626 & 39247 & 314331 \\ \hline
The Coca-Cola Company & 25048 & 14 & 834 & 23544 & 210753 \\ \hline
The Goldman Sachs Group, Inc. & 136787 & 14 & 653 & 66379 & 597153 \\ \hline
The Home Depot, Inc. & 20597 & 16 & 888 & 17242 & 161742 \\ \hline
The Procter \& Gamble Company & 10133 & 17 & 1013 & 8295 & 63014 \\ \hline
The Travelers Companies, Inc. & 4440 & 16 & 953 & 4660 & 46782 \\ \hline
The Walt Disney Company & 51512 & 14 & 711 & 38655 & 319070 \\ \hline
UnitedHealth Group Incorporated & 16389 & 15 & 935 & 13237 & 135272 \\ \hline
Verizon Communications Inc. & 22475 & 14 & 774 & 21746 & 161878 \\ \hline
Visa Inc. & 25452 & 14 & 872 & 20903 & 197226 \\ \hline
Walgreens Boots Alliance, Inc. & 9817 & 15 & 776 & 11964 & 99190 \\ \hline
All companies & 1332446 & 14 & 799 & 275341 & 2726519 \\ \hline
\end{tabular}
\caption{\label{tb:corpus}Corpus statistics for all companies, including the total number of articles, average title and text lengths (in words), and vocabulary sizes measured over the duration of the backtesting period.}
\end{table}

\begin{figure*}[!h]
    \centering
    \includegraphics[width=1\linewidth]{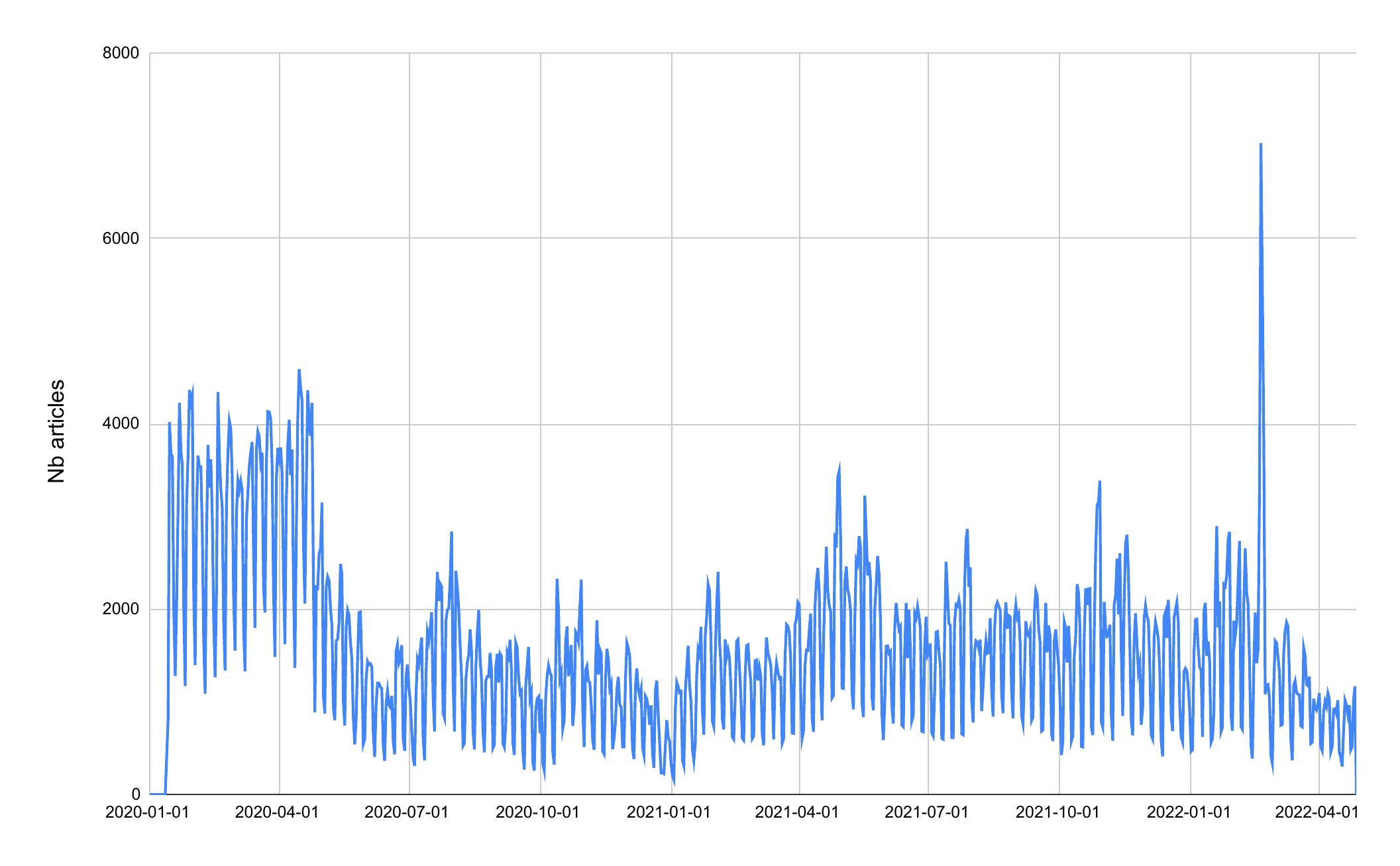}
    \caption{ Volume of published articles over the backtesting period.}
    \label{fig:article-volume}
\end{figure*}

Figure~\ref{fig:article-volume} presents the distribution of published article volumes across the backtesting period, highlighting temporal trends in news coverage. Notably, one of the peaks in article volume observed in February 2022 corresponds to the onset of the Russo-Ukrainian War, which triggered a surge in media attention.

\section{Appendix: Sentiment model training}
\label{sc:train_models}

This section outlines the training methodology employed for developing the sentiment analysis classification and regression models. The FinBERT model, accessible at \url{https://huggingface.co/ProsusAI/finbert}, served as the foundational pre-trained model for our classification efforts.

For the DualGCN model, we created an aggregated financial dataset composed of 4,846 sentences from the Financial Phrase Bank v.1.0 and 1,633 news headlines from the SemEval-2017 task \#5 with 80\% randomly selected samples, leaving the other 20\% as the development dataset.
News headlines from the SemEval-2017 task \#5 contain a sentiment score which is comprised between $-1$ and $+1$, given that our objective is to categorize sentiments with a discrete approach, we transformed the closed interval $[-1, +1]$ into \textit{negative} $[-1.0, -0.1)$, \textit{neutral} $[-0.1, +0.1]$, and \textit{positive} $(+0.1, +1.0]$.
To obtain an accurate feature extraction and in-domain meaning, we used FinBERT for text tokenization and representation.

For the RoBERTa+Transfs model, we followed the same training procedure as described in \cite{roberta-transformer}. Since the model outputs regression values in the range $[-1,+1]$, cosine similarity was used to assess the closeness between the predicted values and the gold standard, where higher scores (closer to 1) indicate better alignment. The model reproduced the results reported in the original study, achieving a cosine similarity score of $0.848$.

To benchmark the performance of FinBERT and DualGCN, we conducted evaluations on the classification task of the Financial PhraseBank dataset. The results, as depicted in Table~\ref{tb:finbertvsdualgnc}, highlight the superior performance of DualGCN, exhibiting higher accuracy and F1-score compared to FinBERT.

\begin{table}[h]
\centering
\begin{tabular}{|l|c|c|c|c|c|}
\hline
\textbf{System}  & \textbf{Loss} & \textbf{Acc}  & \textbf{Precision}  & \textbf{Recall} & \textbf{F1}    \\\hline
Finbert     & 0.37    & 0.86   & -  & - & 0.84 \\\hline
DualGCN     & \textbf{0.25}    & \textbf{0.94}   & 0.93  & 0.95 & \textbf{0.93}  \\\hline
\end{tabular}
\caption{Performance of Finbert and DualGCN methods on Financial PhraseBank dataset classification task. The best results are in bold.}
\label{tb:finbertvsdualgnc}
\end{table}

\section{Appendix : Evaluation metrics}
\label{sc:annex_metrics}

In the evaluation of financial backtesting, several key performance metrics are of significant relevance, providing valuable information on the effectiveness and risk associated with investment strategies. In this work, we selected metrics such as annual, annual compound, and annual cumulative returns; Calmar, Sharpe, and Sortino ratios; max drawdown (MDD); annual volatility; and 95\% daily value at risk (VaR). These metrics collectively offer a holistic view of a strategy's performance, risk, and resilience. Investors rely on these measures to assess the profitability and risk associated with their investments.

\paragraph{Annual Return} It quantifies the percentage increase or decrease in the value of an investment over one year. It takes into account both the capital appreciation (or depreciation) of the investment and any income generated from it, such as dividends, interest, or rental income. It is calculated as 

\begin{equation}
    \textrm{Annual Return} = \frac{V_e - V_b + V_i}{V_b},
\end{equation}

\noindent where $V_e$ and $V_b$ correspond to the values at the ending and beginning of the one-year period, while $V_i$ refers to any income generated from the investment during the year.

\paragraph{Annual Compound Return} It is a measure that calculates the average annual growth rate of an investment over a specific period, considering the effects of compounding. It takes into account the fact that investment returns are often reinvested, leading to compounding growth. This means that not only is the initial investment growing, but the returns generated by the investment are also contributing to further growth.
It is calculated as 

\begin{equation}
    \textrm{Annual Compound Return} = \left( \left( \frac{V_e}{V_b} \right)  ^\frac{1}{n} -1 \right) \times 100 ,
\end{equation}

\noindent where $V_e$ and $V_b$ refer to the value of an investment at the end and the beginning of the period, and $n$, is the number of years over which the investment has grown.

\paragraph{Annual Cumulative Return} It represents the total change in the value of an investment or portfolio over a specified period independently of the amount of time involved. It shows how much the investment's value has grown or declined from the beginning of the period to the end, considering both capital appreciation and any income generated, such as dividends, interest, or other distributions.
It is calculated by taking the difference between the ending value ($V_e$) and the beginning value ($V_b$) of the investment and then expressing that change as a percentage of the beginning value as follows:

\begin{equation}
    \textrm{Annual Cumulative Return} = \left( \frac{V_e - V_b}{V_b} \right) \times 100 .
\end{equation}

\paragraph{Max Drawdown (MDD)} It measures the most significant peak-to-trough decline in the value of an investment or portfolio over a specific period. MDD quantifies the worst loss an investor could have experienced if they had invested at the peak and sold at the trough during that period. It is calculated as 

\begin{equation}
     \textrm{MDD} = \frac{V_t - V_p}{V_p} \times 100,
\end{equation}

\noindent where $V_t$ and $V_p$ are the lowest and highest points or values the investment reaches during the specified period, respectively.

\paragraph{Calmar Ratio} It is a risk-adjusted performance measure that evaluates the return of an investment in relation to its MDD. It is specifically designed to assess the risk-adjusted return of investments or portfolios, taking into account the potential losses incurred during market downturns. It is calculated as

\begin{equation}
    \textrm{Calmar Ratio} = \frac{R_p - R_f}{\textrm{MDD}},
\end{equation}

\noindent where $R_p$ refers to the expected or average return of the investment and $R_f$ is the risk-free rate of return.

\paragraph{Sharpe Ratio} It represents the risk-adjusted return of an investment or portfolio. It takes into account both the investment's potential return and the level of risk it carries.
The Sharpe ratio \citep{sharpe1998sharpe} is calculated by subtracting the risk-free rate of return ($R_f$) from the expected or average return of the investment ($R_p$), and then dividing the result by the standard deviation of the investment returns ($\sigma_p$) as follows:

\begin{equation}
    \textrm{Sharpe Ratio} = \frac{R_p - R_f}{\sigma_p}
    \label{eq:shape_r}
\end{equation}

\paragraph{Sortino Ratio} It is a risk-adjusted performance measure to evaluate the return of an investment in relation to its downside risk, specifically considering only the volatility associated with negative price movements or losses. The Sortino ratio \citep{sortino1994performance} takes an asset or portfolio's average return ($R_p$), subtracts the risk-free rate ($r_f$), and then divides that amount by the asset's downside standard deviation ($\sigma_d$). Unlike the more common Sharpe ratio, which uses an investment's total volatility (both upward and downward), the Sortino ratio focuses only on the volatility of negative returns. It is defined as follows:

\begin{equation}
    \textrm{Sortino Ratio} = \frac{R_p - r_f}{\sigma_d}
\end{equation}

\paragraph{Annual Volatility} It measures the dispersion of an investment's returns from its average return over one year. It quantifies the degree of fluctuation or variability in the price or value of an investment. Since volatility describes changes over a specific period, it is calculated by multiplying the standard deviation of returns ($\sigma$) by the square root of the number of periods in a year ($T$) as follows:

\begin{equation}
    \textrm{Annual Volatility} = \sigma \times \sqrt{T}
\end{equation}

\paragraph{95\% Daily Value at Risk (VaR)} It is a risk management measure that quantifies the potential loss an investment or portfolio might experience over a specified time horizon at a given confidence level. The 95\% daily VaR, in particular, is a specific instance of VaR where the confidence level is set at 95\% and the time horizon is one trading day.

\end{document}